\documentclass[runningheads]{llncs}

\usepackage[T1]{fontenc}

\usepackage{graphicx}

\usepackage{amsmath}
\usepackage{booktabs}
\usepackage{subcaption}
\usepackage{amssymb}

\begin{document}

\title{Approximating Human Strategic Reasoning with LLM-Enhanced Recursive Reasoners Leveraging Multi-agent Hypergames}

\titlerunning{LLM Reasoners in Multi-level Hypergames}

\author{Vince Trencsenyi\inst{1} \and
Agnieszka Menselt\inst{1} \and
Kostas Stathis\inst{1}}

\authorrunning{V. Trencsenyi et al.}

\institute{Royal Holloway University of London, Egham, Surrey, UK\\
\email{\{vince.trencsenyi, agnieszka.mensfelt, kostas.stathis\}@rhul.ac.uk}}

\maketitle

\begin{abstract}
LLM-driven multi-agent-based simulations have been gaining traction with applications in game-theoretic and social simulations. While most implementations seek to exploit or evaluate LLM-agentic reasoning, they often do so with a weak notion of agency and simplified architectures.
We implement a role-based multi-agent strategic interaction framework tailored to sophisticated recursive reasoners, providing the means for systematic in-depth development and evaluation of strategic reasoning. Our game environment is governed by the umpire responsible for facilitating games, from matchmaking through move validation to environment management. Players incorporate state-of-the-art LLMs in their decision mechanism, relying on a formal hypergame-based model of hierarchical beliefs. We use one-shot, 2-player beauty contests to evaluate the recursive reasoning capabilities of the latest LLMs, providing a comparison to an established baseline model from economics and data from human experiments. Furthermore, we introduce the foundations of an alternative semantic measure of reasoning to the k-level theory. Our experiments show that artificial reasoners can outperform the baseline model in terms of both approximating human behaviour and reaching the optimal solution.

\end{abstract}

\section{Introduction}

Multi-agent systems provide environments for individual-based modelling and simulations~\cite{michel2018multi}. Game theory and multi-agent-based simulation (MABS) have established a mutually beneficial relationship: MAS leverages game-theoretic interaction models and strategic tools~\cite{parsons2002game}, and game theory relies on multi-agent-based social simulations to investigate strategic decision-making~\cite{sun2006cognition}. Large language models (LLMs) have received particular interest in their potential to simulate human-like reasoning and decision-making, and MABS frameworks are used to evaluate LLM capabilities in game-theoretic environments~\cite{guo2024economicsarenalargelanguage}. Traditional approaches often rely on simplified agent frameworks implemented with a weaker concept of agency~\cite{Wooldridge_Jennings_1995}, which may impose limitations on the system's adaptability and the agent's reasoning sophistication. In contrast, our approach involves a stronger, modular agent concept enabling a decoupled investigation of reasoning processes. Our agents are hosted in a role-based multi-agent framework governed by an umpire who manages game environments and facilitates agent interactions.

We focus on beauty contest games, a well-established concept for studying recursive reasoning~\cite{camerer2004cognitive}. We evaluate the reasoning capabilities of LLM-enhanced agents, integrating recursive reasoning via a formal hypergame representation. Our experiments compare the performance of LLM-enhanced agents against both a baseline economic model and human data, providing insights into the models' ability to approximate human strategic behaviour. Furthermore, we introduce a self-evaluation method $\kappa$, a revised measure of reasoning depth that complements traditional k-level theory, offering a more nuanced understanding of reasoning sophistication.

Our key contributions include:
\begin{itemize}
    \item A flexible multi-agent-based simulation platform capable of hosting a wide array of reasoners, offering a detailed view of reasoning processes;
    \item LLM-enhanced agents that leverage a hypergame-based model for recursive reasoning;
    \item Introduction of $\kappa$, a complementary measure to k-level reasoning;
    \item Experiments comparing our LLM-based reasoners to the baseline model and human data.
\end{itemize}

Our results suggest that artificial reasoners can benefit from the expanded architectural complexity and can not only match but potentially outperform baseline models in both approximating human behaviour and achieving optimal solutions in strategic settings.

\section{Background}

\subsection{Game Theory}

Game theory provides a mathematical framework for analyzing decision-making in multi-agent -- human or artificial -- strategic interactions~\cite{Osborne2004}. A game is formally defined by its players, their available strategies, and utility functions that evaluate the players' outcomes~\cite{Rasmusen2006introtogametheory}. Beauty contest games (BCGs) provide an experimental testbed for iterative reasoning, where players have to guess a number which they believe will be the closest to the mean of all guesses weighted by a parameter $p$~\cite{camerer2004cognitive}. BCGs are a popular choice for studies concerning the k-level theory, as optimal play requires thinking about others' thoughts: given the range $[0,100]$ and $p=\frac{2}{3}$, 0-level players pick 50 and level-k thinkers choose $50\frac{2}{3}^k$. Experimental evidence consistently shows that most human players exhibit $1^{st}$ or $2^{nd}$-level reasoning, with few advancing beyond level 3~\cite{nagel95,duffy1997robustness,bosch2002one}.

Hypergames extend standard game theory by modelling individual player perspectives, allowing hypergame models to capture misaligned perceptions~\cite{Bennett1977}. Multi-level hypergames provide a formalized model of hierarchical games representing nested beliefs~\cite{Wang1988}. A third-level hypergame between players $i,j$, is a composite structure of lower-level hypergames representing individual players' perspectives: $H^3=\{H^2_i,H^2_j\}$, where $H^2_i=\{H^1_{ii},H^1_{ji}\}$ and $H^1_{ji}=\{H^0_{iji},H^0_{jji}\}$. Finally, $H^0_{iji}=G_{iji}$ is $i$'s perceptual game defining player $i$'s belief of $j$'s belief of $i$'s perspective of $G$. This theoretical framework provides the foundation for studying how agents engage in recursive strategic reasoning and form hierarchical beliefs about others' decision-making processes.

\subsection{Language Models}

Large language models (LLMs) are sophisticated neural networks -- usually based on the transformers architecture~\cite{gillioz2020overview} and a pre-trained model on a vast amount of data~\cite{qiu2020pre} -- that target natural language processing applications~\cite{zhao2023survey}. Chain-of-thought (CoT) prompting is a technique that improves LLMs reasoning capabilities by having LLMs decompose complex tasks into smaller problems~\cite{wei2023chainofthoughtpromptingelicitsreasoning}. Such prompts can be demonstrative examples showcasing what the expected response may pertain to and/or descriptive instructions that guide the model on reaching the expected response~\cite{yu2023towards}.

\subsubsection{Claude 3.5}
Claude 3.5 is a family of state-of-the-art LLM models from Anthropic, supporting multimodal applications~\cite{claude35}. In this work, we evaluate two Claude models: Sonnet offers large context windows and advanced analytical capabilities, suitable for complex tasks and process automation; Haiku is a smaller, cost-efficient, fast model targeting interactive and sub-agent tasks.

\subsubsection{GPT-4}
GPT-4 is OpenAI's SOTA system supporting multimodal input and output and a long context window arming models with a broad general knowledge and advanced problem-solving abilities~\cite{openai2024gpt4technicalreport}. We implement two models: GPT-4o possesses a generally high reasoning performance across various benchmarks; 4o-mini is a lightweight variant for resource-constrained environments.

\subsection{Recursive Reasoning}

We describe recursive reasoning as an agent's ability to reason about each other's physical and cognitive states~\cite{wen2019probabilistic}. The cognitive hierarchy model introduced the concept of k-level thinkers, proposing that players engage in different levels of strategic thinking, where players on level $k$ best respond to $k-1$ level strategies, assuming every other player must be at most at level $k-1$~\cite{camerer2004cognitive}. Epistemic game theory is a branch of game theory which provides a formal mathematical framework for representing and operating with belief hierarchies~\cite{dekel2015epistemic}. Hypergame theory aims to analyse conflict under asymmetric information and misaligned perceptions~\cite{Bennett1980}, providing a game-oriented representation of sequential beliefs.

\section{Multi-agent Simulation via Centralized Hypergames}

Our simulations are materialized in game environments composed of and hosting an umpire, a set of players, and a set of hypergames. We revise hierarchical hypergames from~\cite{Wang1988} to integrate 2-player BCGs as individual perceptual games capturing the agent's beliefs and reasoning level.

\subsection{Perceptual Beauty Contest Games}

In our framework, we define BCGs formally as $G=(N,A,U,\Psi)$, where $N={i,j}$ is the set of two players, $A = A_i \times A_j$ is the action space, where $A_i, A_j \subseteq \mathbb{Z}$ represent the available actions for players $i$ and $j$ respectively, $U: A \rightarrow \mathbb{R}^2$ is the utility function, where for each player:
$U_i(a_i,a_j) = -|a_i - p \cdot \mu|$, with $\mu = \frac{a_i + a_j}{2}$ and $p$ as the BCG's scalar parameter and $\Psi=(\psi_1=\sigma,\psi_2,\ldots,\psi_\kappa)$ is an ordered sequence of $\kappa$ number of perspectives, where the first component $\psi_1$ denotes the interpreter (creator) of the game perspective, $\sigma$. To help position player beliefs in our BCGs in the context of belief hierarchies, we suggest simplifying assumptions as follows. Given a $\kappa=2$ level reasoner $i$, $\beta_i(\beta_j(\beta_k))$~\cite{dekel2015epistemic} corresponds to $i$'s beliefs about player $j$'s beliefs about player $k$'s reasoning. Then we assume $\beta_i(\beta_j(\beta_k)) \cong G_{ijk}$, where $G_{ijk}$ is the perceptual game capturing the same beliefs, with subscripts denoting the sequential order of perspectives: $i$'s reconstruction of $G$ based on his beliefs about $j$'s beliefs of $k$'s perspective. However, the rigorous analysis of the proposed relationship is beyond the scope of this work. Finally, the set of individual perceptual games invokes the hypergame $H^{\kappa}$, capturing both players' beliefs, where at least one of the players exhibits the highest level of reasoning $\kappa$.

\subsection{Recursive Reasoners}

\begin{figure}[tb]
\centering
\begin{subfigure}{0.35\linewidth}
  \centering
  \includegraphics[width=\linewidth]{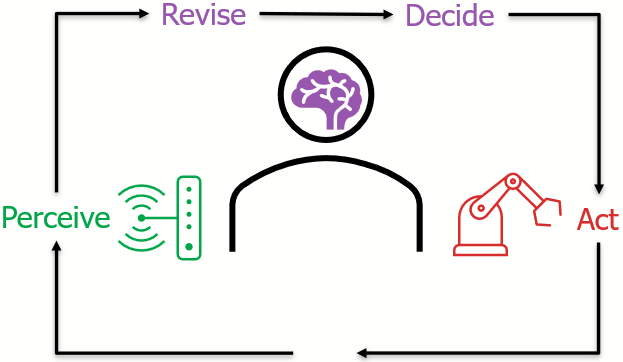}
  \caption{The modular agent relies on its sensor (green) to perceive its environment, uses its mind (purple) to revise the perception and decide which action to take, and acts upon the environment via its effector (red).}
  \label{fig:agent-cycle-a}
\end{subfigure}
\hfill
\begin{subfigure}{0.625\linewidth}
  \centering
  \includegraphics[width=\linewidth]{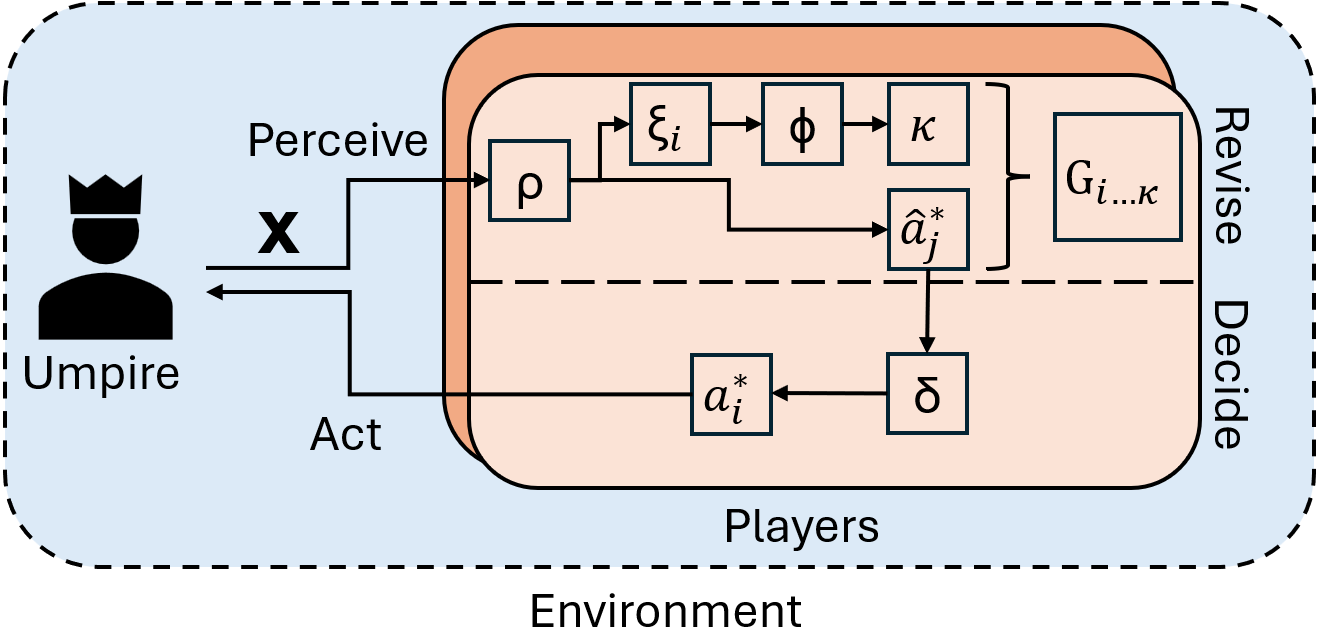}
  \caption{Upon perception of a game request with textual description $x$, the player revises $x$, generating a reasoning $\xi_i$ and producing an expected opponent choice $\hat{a_{j}}$. After $\phi(\xi_i)$ derives $\kappa$, the perceptual game $G_{i\ldots \kappa}$ is established. $\delta(\hat{a_{j}})=a^*_i$ is what $i$ decides and acts on.}
  \label{fig:agent-cycle-b}
\end{subfigure}
\caption{Overview of centralised MAS framework.}
\label{fig:agent-cycle}
\end{figure}

Agents in our framework are implemented following the standard intelligent agent architecture~\cite{russell1995intelligent} and inspired by the Observe-Orient-Decide-Act decision-making model~\cite{bryant2006rethinking,OODA_agents} -- as shown on Figure~\ref{fig:agent-cycle-a}. An iteration of the game environment comprises the umpire's matchmaking activities and the players' reasoning processes. Given the space of natural language game descriptions $X$, the umpire $\upsilon$ sends game requests to each pair of players to participate in $G^*=(\{i,j\},\{A_i,A_j\},U,(\upsilon))$. In the facilitated games, the umpire is a passive participant -- a pseudoplayer~\cite{Rasmusen2006introtogametheory}.

Each player's reasoning processes are then decoupled and defined as follows:
\begin{itemize}
    \item The player's revision module -- Figure~\ref{fig:agent-cycle-b} -- processes the game description, reasons about what the opponent's move could be, and constructs a perceptual game. These steps are integrated into an interpretation function $I:G \times N \rightarrow G'$, that creates an instance of $G$ reflecting the player's beliefs:
    \begin{itemize}
        \item $\rho: X \rightarrow \Xi, \mathbb{R}$ is the reasoning function: $\rho(x)=\xi_i,\hat{a_{j}}$ where $\xi_i \in \Xi$ is $i$'s natural language reasoning based on game description $x \in X$ on what the opponent $j$'s guess $\hat{a_{j}}$ is expected to be;
        \item $\phi: \Xi \rightarrow \mathbb{N}$ is the reasoning analysis function: $\phi(\xi_i) = \kappa$, where $\kappa$ is the player's estimated reasoning level based on the number of nested beliefs present in the reasoning $\xi_i$;
        \item $i$'s perceptual game is then: $G_{i\ldots \kappa}=(i,\{i,j\},A,U,\Psi=(\psi_i,\ldots,\psi_\kappa))$.
    \end{itemize}
    \item The decision module selects an action based on the revised expectation:
    \begin{itemize}
        \item $\delta: \mathbb{R} \rightarrow A_i$ is the decision function selecting $i$'s preferred action $a^*_i$ based on the expected opponent guess: $\delta(\hat{a_j}) = a^*_i$;
    \end{itemize}
\end{itemize}

While k-levels are derived from players' numerical choices, $\kappa$ provides an alternative measure based on the explicit reasoning steps we observe in players' natural language explanations. These reasoning steps are represented as perspectives $\psi$ in the perceptual game $G$, allowing us to analyze the depth of strategic thinking through players' own articulation of their decision process.

\subsection{Model Prompting Design}

We integrate CoT prompting into our multi-step reasoning process across the agent's \textit{revise} and \textit{decide} phases. During \textit{revise}, the LLM mind processes information through sequential steps: first reasoning about the situation to predict opponent guesses, then analyzing the reasoning depth to determine $\kappa$, resulting in the $\kappa^{th}$ order perceptual game. The mind's decision function then interprets this processed information to derive the final guess, completing the multi-step reasoning process. LLM prompts follow a modular structure: (1) optional agent profile for context-specific reasoning~\cite{mao2024alympicsllmagentsmeet}, (2) role and task definition, (3) game/task specific request, and (4) implementation-specific requirements (e.g., \textit{``surround the chosen number in curly brackets: {n}''}).

\subsection{Baseline Model}
The self-tuning experience weighted attraction model (EWA) is a benchmark model for reproducing human-like strategic interaction in game-theoretic experiments~\cite{ho2007self}.

\begin{equation}
A^j_i(t)=\frac{\phi \cdot N(t-1) \cdot A^j_i(t-1)+[\delta+(1-\delta) \cdot I(s^j_i,s_i(t))] \cdot \pi_i(s^j_i,s_{-i}(t))}{N(t-1) \cdot \phi \cdot (1-\kappa)}
\end{equation}

Equation 1 defines player $i$'s associated attention to strategy $j$ at time (or round) $t$, where
\begin{itemize}
    \item $\pi_i(s^j_i,s_{-i}(t))$ denotes $i's$ payoff for choosing strategy $j$ against $s_{-i}$ at time $t$;
    \item $N$ is the experience weight $N(t) = (1-\kappa) \cdot \phi \cdot N(t-1)+1$~\cite{CAMERER2002137};
    \item $\phi$ is the change detector function $\phi(t)=1-\frac{1}{2}S_i(t)$;
    \item the surprise index is denoted by $S_i(t)=\sum^{m_{-i}}_{k=1}{(h^k_i(t)-r^k_i(t))^2}$ with\\ $r^k_i(t)=I(s^k_{-i},s_{-i}(t))$ and $h^k_i(t)=\frac{\sum^{t}_{\tau=1}{I(s^k_{-i},s_{-i}(\tau))}}{t}$;
    \item $\delta$ is the weight to foregone payoffs: $\delta_{ij}(t)= \begin{cases}
        1 & \text{if } \pi_i(s^j_i,s_{-i}(t)) \geq \pi_i(t),\\
        0 & \text{otherwise.}
    \end{cases}$;
    \item and $I(x,y)$ is an index function returning  $0$ if $x=y$ and $1$ otherwise.
\end{itemize}
Let $t=0$ denote the agent's initial state. We then define the initial attraction $A^j_i(0)$ according to the cognitive hierarchy model and the Poission distribution function $P(k) = \frac{e^{-\tau}\tau^k}{k!}$~\cite{camerer2004cognitive} and set $N(0)=1$. The self-tuning EWA model was trained and tested on a 7-player beauty contest with $p=0.7$ and $p=0.9$~\cite{ho1998iterated}, for which Ho et al. determined $\lambda=2.39$ for the sensitivity of the response function and $\tau=1.5$ for deriving the first-period plays via the CHM-derived Poisson distribution. We adopt these parameters in our experiments without further tuning. Finally, agents choose an action according to $P^i_j(t+1) = \frac{e^{\lambda \cdot A^j_i(t)}}{\sum^{m_i}_{k=1}{e^{\lambda \cdot A^k_i(t)}}}$.

\section{Experiments}

We use beauty contest games as the test bench for evaluating our agents' recursive reasoning capabilities, as guessing games are closely associated with k-level reasoning~\cite{camerer2004cognitive}. In this context, $k$ level reasoners best respond to the $k-1$ level players' choices and at $k=0$ players choose randomly. In order to classify players by their reasoning, we adopt the guess-based conversion approach presented in~\cite{nagel95}. We assume level 0 reasoners choose 50; then we set each $k^{th}$ reasoning level at $50p^k$.
More specifically, we use 2-player BCGs to conduct our evaluation. In $n$-player BCGs, reasoners eliminate weakly dominated strategies iteratively -- which process can be extended to infinity -- until the theoretical solution of everyone choosing 0 is reached~\cite{camerer2004cognitive}. The two-player game provides a simpler solution concept, where the smaller number wins~\cite{nagel08}. Zero is a weakly dominated strategy that always wins, which, in theory, significantly simplifies the iterative reasoning process.

We evaluate our agents on the experiment from~\cite{nagel08}, involving 132 student participants, pooled from first-year students with no prior game-theoretic training nor existing familiarity with beauty contests and 130 professionals with extensive game-theoretic domain knowledge.

In the first instance, we replayed the 2-player beauty contests using the original experiment design, simulating 25 independent rounds with each LLM. Additionally, we conducted 60 rounds with pairs of agents using the EWA model as the benchmark. Artificial agents were provided with a description worded similarly to what human participants would be given -- however, obfuscated to mitigate reliance on game-theoretic analyses in the models' training data --, without any explicit instructions on their reasoning and decision-making. Providing a persona description as context pushes the LLM to behave in the desired way by guiding the reasoning process according to the profile specification~\cite{mao2024alympicsllmagentsmeet}. Similarly, in the second set of experiments, we expand our prompting mechanism with an agent profile, allowing us to specify the level of domain knowledge the agent should use for its reasoning. Following the original group descriptions from~\cite{nagel08}, we ran 15-15 games with the specifications of ``first years students with no game-theoretic knowledge'' and ``professors with expert domain knowledge in game theory'' with Claude 3.5 Haiku and GPT-4o -- chosen based on their proximity to the human performance and the optimal strategy, 0.

\begin{figure}[htb]
    \centering
    \includegraphics[width=\linewidth]{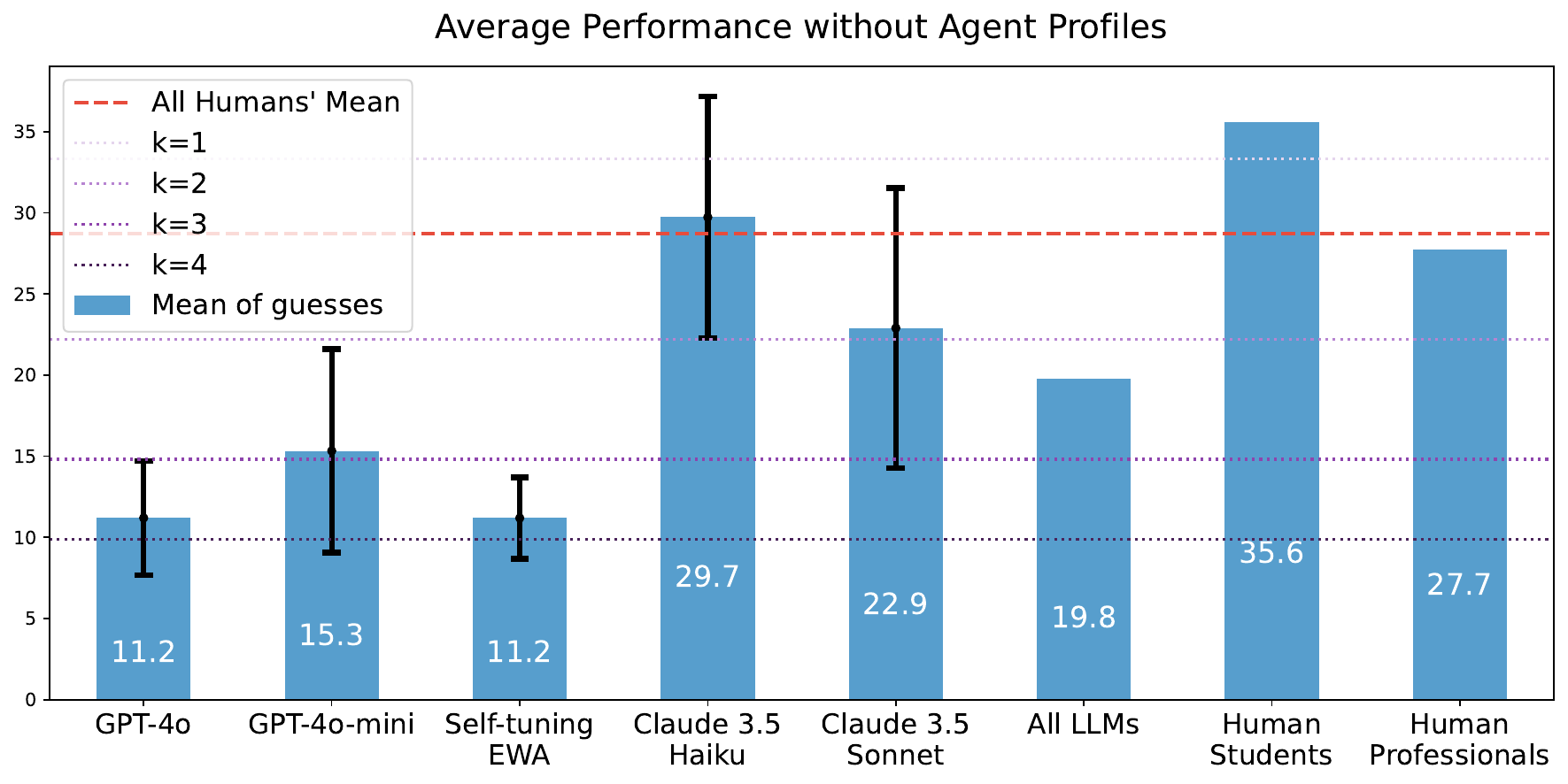}
    \caption{Per model means, standard deviations and estimated $k$-levels. Human data for standard deviation was unavailable.}
    \label{fig:p2-means}
\end{figure}

\begin{figure}[htb]
    \centering
    \includegraphics[width=\linewidth]{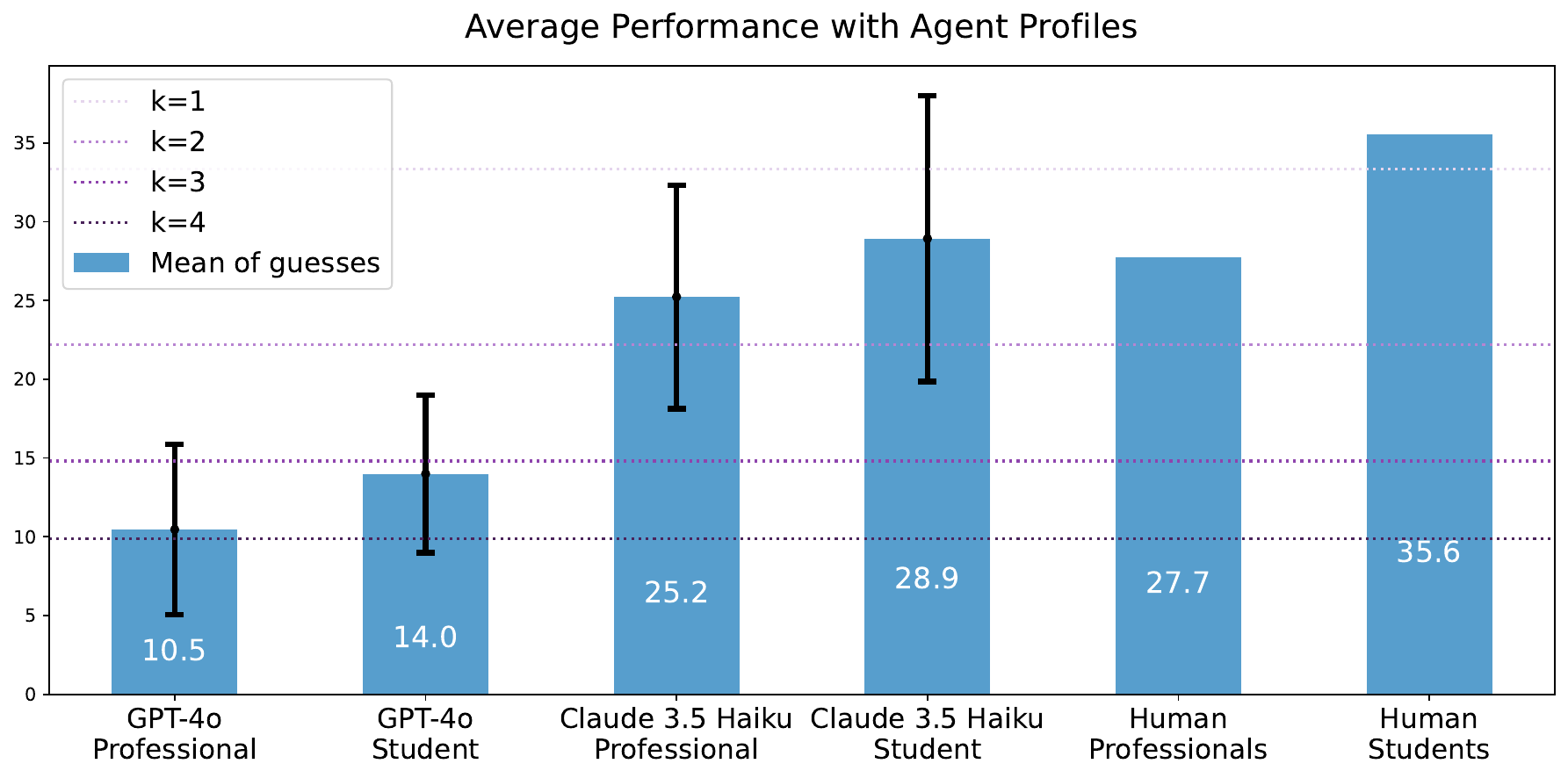}
    \caption{Per model means (LLMs with profiles), standard deviations and estimated $k$-levels. Human data for standard deviation was unavailable.}
    \label{fig:p2-means-profiled}
\end{figure}

All models outperformed human participants significantly, except for Claude 3.5 Haiku, which approximates the results of the professionals and the joint human player population -- Figure~\ref{fig:p2-means}. On the other hand, Figure~\ref{fig:p2-means-profiled} depicts the results of the second experiment, where player profiles are successfully integrated into both agents' reasoning processes. Similar to the human groups with different levels of domain expertise, the models acting as students performed noticeably poorer than the models that were described as professionals. However, while $9.85\%$ of human students and $36.02\%$ of human professionals managed to reach 0, none of the artificial reasoners -- nor the baseline model -- chose the optimal strategy in either experiment. Most studies concentrate on BCGs involving more than 2 players, where 0 is not an obvious solution to the game. This bias is likely inherently present in the LLMs training data, potentially tainting the information used to generate reasoning in the 2-player BCG.

K-level reasoning provides another metric for comparing human and artificial agents' thought processes. Table~\ref{tab:klevels} reinterprets the mean results through k-level classifications, showing that while humans exhibit $1^{st}$-level reasoning, only Claude 3.5 Haiku produced comparable results. Other models performed at the $2^{nd}$ and $3^{rd}$ levels, with EWA and GPT-4o approaching level 4 reasoning. Additionally, the ``Mean $\kappa$'' and ``Median $\kappa$'' columns report the LLM agents' reasoning steps involved, corresponding to perspectives in their perceptual games.

The second experiment revealed that the estimated $k$ and $\kappa$ levels are aligned with the expectation that reasoners with domain expertise would outperform and exhibit a higher order of reasoning than non-professionals. While $\kappa$ can provide a semantics-backed estimate of the reasoning steps involved in the process, it does not yet tell us much about the quality of the agent's reasoning. The results suggest that a semantic qualitative reasoning classification may complement the current numerically defined measures.

\begin{table}[htb]
\centering
\begin{tabular}{lcccc}
\toprule
Model/Group & Mean Guess & Mean k-level & Mean $\kappa$ & Median $\kappa$ \\
\midrule
Human Students & 35.57 & 0.84 & N.A. & N.A.\\
Human Professionals & 27.73 & 1.45 & N.A. & N.A. \\
Claude 3.5 Haiku & 29.7 & 1.28 & 2.17 & 2\\
Claude 3.5 Sonnet & 22.9 & 1.92 & 2.17 & 2\\
GPT-4o-mini & 15.3 & 2.92 & 2.57 & 3\\
GPT-4o & 11.2 & 3.69 & 2.27 & 2\\
Self-tuning EWA & 11.2 & 3.69 & N.A. & N.A.\\
\midrule
Claude 3.5 Haiku Student& 28.9 & 1.35 & 1.97 & 2\\
Claude 3.5 Haiku Professional& 25.2 & 1.69 & 2.5 & 2.5\\
GPT-4o Student & 14 & 3.14 & 1.8 & 2\\
GPT-4o Professional& 10.5 & 3.85 & 2.4 & 3\\
\bottomrule
\end{tabular}
\caption{Mean k-levels are estimated from mean guesses, while mean and median $\kappa$ levels correspond to LLM agents' self-reported reasoning steps.}
\label{tab:klevels}
\end{table}

\section{Related Work}

The recent emergence of LLM agents~\cite{wang2024survey} and LLM-MAS~\cite{li2024survey} has influenced social simulations. Multi-agent strategic interactions are used to evaluate LLM reasoning~\cite{mensfelt2024logicenhancedlanguagemodelagents,duan2024gtbench}, and LLM-based simulations are leveraged for empirical investigations on strategic behaviours~\cite{mao2024alympicsllmagentsmeet,mensfelt2024autoformalizing}. Recent work on LLM-driven reasoning in BCGs focuses on studying agent rationality and reasoning levels~\cite{zhang2024klevelreasoningestablishinghigher,lu2024strategicinteractionslargelanguage,guo2024economicsarenalargelanguage}. Centralized role-based architectures rely on an umpire~\cite{STATHIS1998401} or a game manager~\cite{gamemaster-gameplayerMAS2007} to coordinate game-based interactions -- such as artificial trading~\cite{tradingagentcompetition} or utility markets~\cite{master-slave-energymanagement,masterslavegames2024}.
Hypergame theory, while effective for analyzing complex conflicts post-hoc~\cite{Bennett1980}, has seen limited practical application~\cite{BENNETT1981shippingcrisis} due to challenges in automation. Most agent-based implementations only borrow conceptually~\cite{dharmadhikari2021hypergamepathplanning,Aitchison2021deceptionhypergames}, with few examples of full integration~\cite{Kahn2021thesis,TANG2024networkattackdefense}.

While prior approaches typically implement a looser agentic concept -- a weaker notion of agency~\cite{Wooldridge_Jennings_1995} -- potentially constraining the system's flexibility and bounding LLMs' reasoning~\cite{wang2025limitsllmbasedhumansimulation}, our framework offers enhanced flexibility and depth. Our conceptually elaborate multi-agent architecture enables a more nuanced evaluation process, facilitating a systematic review of LLM reasoning and automating the generation of hypergames. This approach provides valuable insights into LLMs' capabilities for recursive reasoning, particularly their ability to form beliefs about beliefs and develop a theory of mind.

\section{Conclusions and Future Work}

We present a MABS framework integrating hypergames for multi-level reasoning in BCGs. Our contributions are threefold.
First, we introduce a flexible multi-agent simulation platform capable of hosting both simple models like the self-tuning EWA and sophisticated multi-step reasoners powered by state-of-the-art LLMs. The platform's architecture emphasizes a strong notion of agency, allowing for systematic development and evaluation of strategic reasoning capabilities. In contrast, existing approaches often employ simplified architectures with weaker notions of agency, which can limit both applicability and evaluation depth.
Second, we introduce $\kappa$ as a complementary measure to the k-level theory for evaluating reasoning. While k-levels are solely derived from players' numerical choices, $\kappa$ represents the sequential perspectives in an agent's reasoning process, providing an alternative measure of reasoning depth that could offer valuable insights alongside numerical k-level analysis. Third, our 2-player BCG experiments demonstrate that our proposed approach can match and outperform the baseline model in approximating human results and reaching the optimal solution. Additionally, our profile-enhanced agents successfully mirrored the performance gap observed between student and professional human players.

The presented system and experiments highlight the potential for extensions. Integrating semantic analysis techniques for evaluating the quality and coherence of agent reasoning processes would allow us to develop $\kappa$ into a deeper qualitative measure of reasoning sophistication. Recreating the BCG experiment with human reasoning data would allow us to investigate (dis)similarities in how humans and LLM agents approach recursive strategic thinking, both in terms of $k$ and $\kappa$ levels. Implementing multi-round and n-player games would enable studying learning processes in artificial and human agents. These extensions would contribute to a more comprehensive understanding of recursive reasoning in artificial -- and potentially human -- agents while advancing the practical applications of LLM-driven multi-agent systems in strategic decision-making scenarios.\\



{\small
\noindent\textbf{Acknowledgments.}
The first author would like to express his gratitude to David Levine for his valuable insights on the EWA model.\\
This work was supported by a Leverhulme Trust International Professorship Grant (LIP-2022-001).\\

\noindent\textbf{Disclosure of Interests.}
The authors have no competing interests to declare that are relevant to the content of this article. 
}
\bibliographystyle{splncs04}
\bibliography{references}
\end{document}